\documentclass{article}

\PassOptionsToPackage{numbers, sort&compress}{natbib}
\usepackage[preprint]{neurips_2026}

\usepackage[utf8]{inputenc} 
\usepackage[T1]{fontenc}    
\usepackage{url}            
\usepackage{booktabs}       
\usepackage{amsfonts}       
\usepackage{nicefrac}       
\usepackage{microtype}      
\usepackage{xcolor}         
\usepackage{graphicx}
\usepackage{amsmath}
\usepackage{amssymb}
\usepackage[hidelinks]{hyperref}

\hypersetup{
    pdftitle={STEPS: Scene Text Editing with Preserved Style Using Diffusion and Contrastive Style Encoding},
    pdfauthor={Nicolas Thiebaut, Nameer Hirschkind, Xiao Yu, Kyle Spence},
    pdfsubject={Scene text editing},
    pdfkeywords={scene text editing, diffusion models, style encoding}
}

\newcommand{\gridimg}[2]{\raisebox{-0.5\height}{\includegraphics[width=#1]{#2}}}
\newcommand{\gridlabel}[1]{\raisebox{-0.5\height}{\rotatebox{90}{#1}}}

\makeatletter
\renewcommand{\@noticestring}{%
    A version of this paper appeared in \emph{Data Science: Foundations and Applications (PAKDD 2026)},
    Lecture Notes in Computer Science, vol.~16618, pp.~473--484, Springer Nature Singapore.
    \href{https://doi.org/10.1007/978-981-92-1947-6_38}{doi:10.1007/978-981-92-1947-6\_38}.%
}
\makeatother

\title{STEPS: Scene Text Editing with Preserved Style\\Using Diffusion and Contrastive Style Encoding}

\author{%
    Nicolas Thiebaut \qquad Nameer Hirschkind \qquad Xiao Yu \qquad Kyle Spence \\
    Roblox \\
    San Mateo, CA, USA \\
    \texttt{nthiebaut@roblox.com} \\
}

\begin{document}

\maketitle

\begin{abstract}
We introduce Scene Text Editing with Preserved Style (STEPS), a novel diffusion model architecture for quality text replacement in images.  Scene Text Editing (STE), also known as Visual Text Editing, consists of changing the textual content in an image while conserving the original style, e.g. font, colors, orientation, background, etc. STEPS advances the state of the art in STE through directed focus on improved style preservation. We introduce a style encoder for visual text that captures style independently of textual content, and a model architecture that combines the style encoder with multiple semantic conditions (target text characters encoding and rendered glyphs).

STEPS achieves superior results to previous STE methods in style preservation, output readability, and subjective quality.
\end{abstract}

\begin{figure}[htb]
    \centering
    {\small
    \setlength{\tabcolsep}{1pt}
    \begin{tabular}{@{}c@{\hspace{4pt}}ccc@{\hspace{8pt}}ccc@{}}
        & \multicolumn{3}{c@{\hspace{8pt}}}{Indonesian Translation} & \multicolumn{3}{c@{}}{Italian Translation} \\[2pt]
        \gridlabel{Input Images} &
        \gridimg{0.15\linewidth}{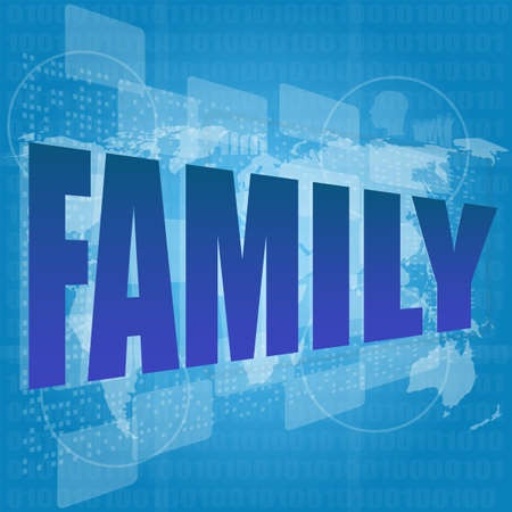} &
        \gridimg{0.15\linewidth}{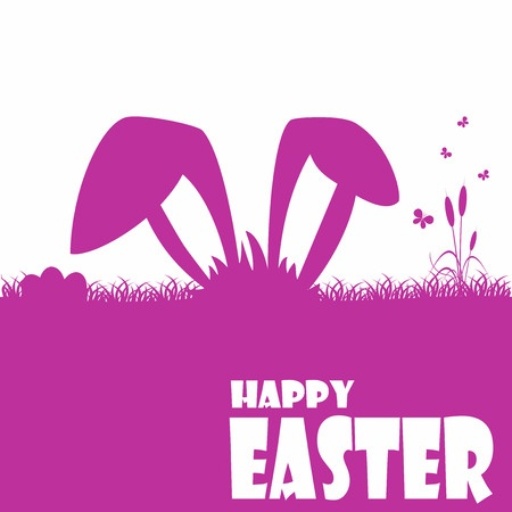} &
        \gridimg{0.15\linewidth}{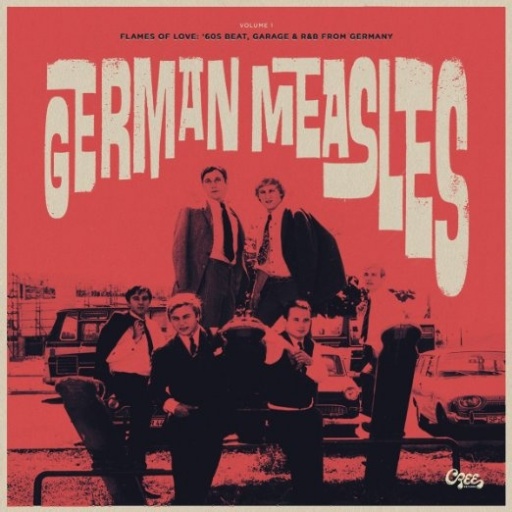} &
        \gridimg{0.15\linewidth}{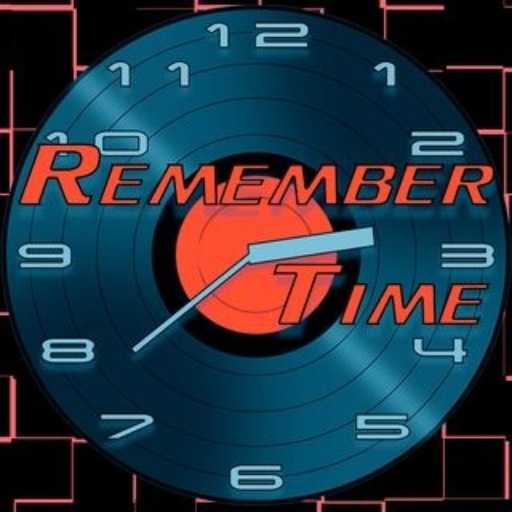} &
        \gridimg{0.15\linewidth}{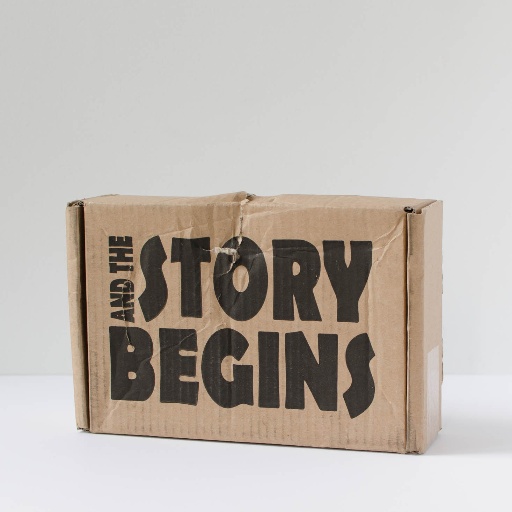} &
        \gridimg{0.15\linewidth}{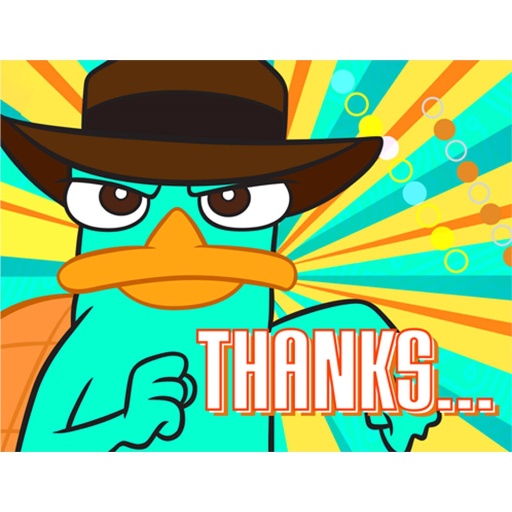} \\[2pt]
        \gridlabel{STEPS Results} &
        \gridimg{0.15\linewidth}{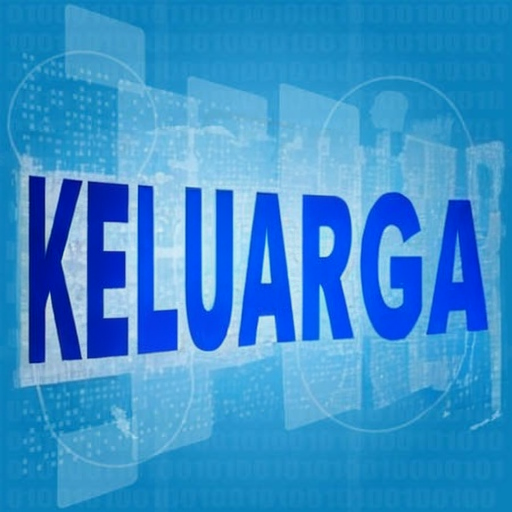} &
        \gridimg{0.15\linewidth}{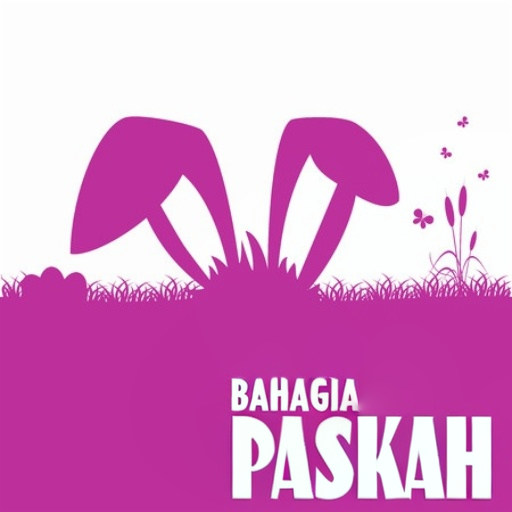} &
        \gridimg{0.15\linewidth}{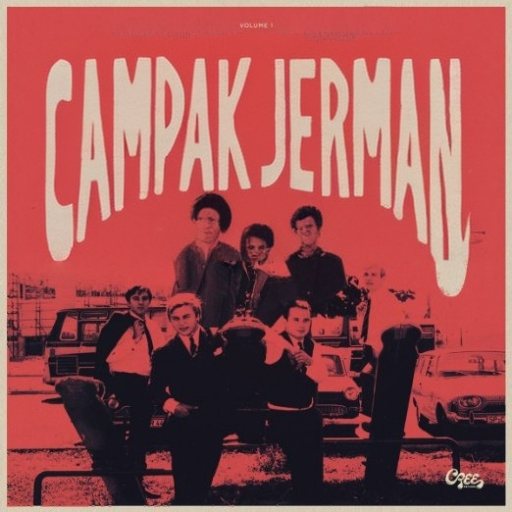} &
        \gridimg{0.15\linewidth}{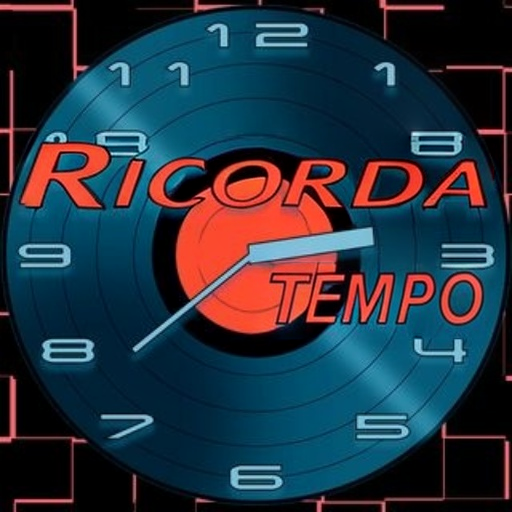} &
        \gridimg{0.15\linewidth}{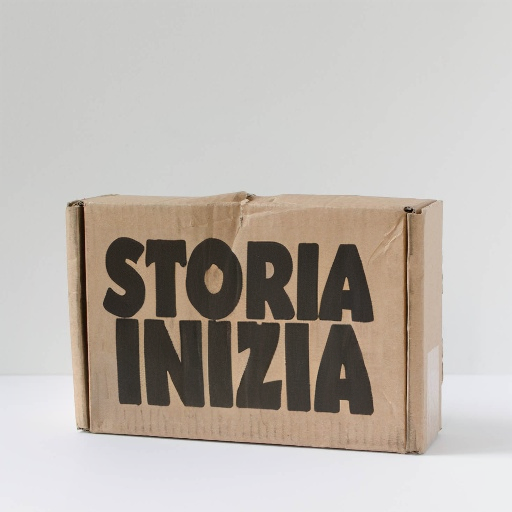} &
        \gridimg{0.15\linewidth}{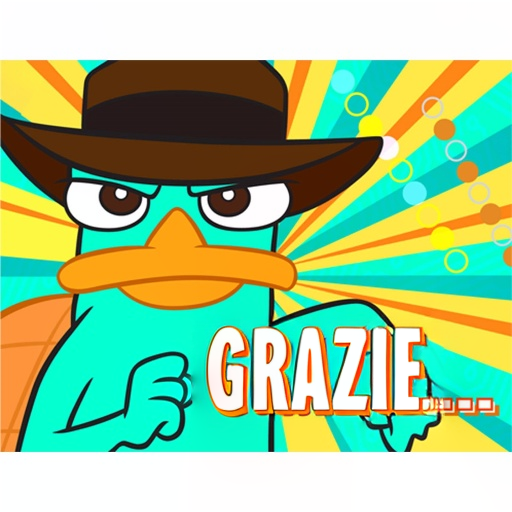} \\
    \end{tabular}}
    \caption{Examples of Scene Text Editing with STEPS applied to natural language translation of visual text in images with arbitrary backgrounds, fonts, and styles. (Top) Input images. (Bottom) translations of visual text into Indonesian (first three) and Italian (last three). STEPS is able to preserve the original text style, plausibly inpaint the disoccluded background, and to accommodate target texts with different lengths than the original.}
    \label{fig:examples}
\end{figure}

\section{Introduction}
\label{sec:intro}

The landscape of image generation and editing has been fundamentally reshaped by recent advances in diffusion models. While these models have established themselves as the state-of-the-art for general synthesis---driven by developments in latent space modeling and improved denoising---the precise rendering of visual text remains a persistent challenge. Generative models frequently exhibit failure modes such as missing character strokes, hallucinating superfluous letters, or failing to strictly adhere to textual prompts.

Within this domain, Scene Text Editing (STE) presents a uniquely constrained problem. Unlike general text generation, a model must not only modify the semantic content but also rigorously preserve the stylistic attributes---font, color, orientation, and background texture---of the original image.

Historically, STE was dominated by Generative Adversarial Networks (GANs) operating at the text-patch level. While approaches like SRNet and MOSTEL established early benchmarks, their reliance on isolated patches prevents the utilization of global image context, often necessitating complex blending steps that result in visual artifacts. Conversely, recent diffusion-based approaches have improved inpainting capabilities but struggle to balance readability with style consistency. We observe that existing techniques \cite{chen_diffute_2023,tuo_anytext_2024,anytext2,ji_diffste_2023,zeng2024textctrl,rsste,kulikov2025floweditinversionfreetextbasedediting} particularly struggle when the target text length deviates significantly from the original, a limitation not fully captured by current benchmarks. Furthermore, state-of-the-art large-scale editing models, such as Flux.1 Kontext and Qwen-Image-Edit, remain computationally demanding and lack the specific fine-grained control required for multilingual scene text replacement.

To address these limitations, we introduce Scene Text Editing with Preserved Style (STEPS), a novel diffusion architecture designed to achieve superior style preservation without compromising output readability.

Our approach bridges the gap between style and content through two key contributions:

\begin{enumerate}
    \item A Novel Style Encoder: We propose a visual text style encoder trained on a unique heuristic: that spatially close visual text tends to share similar stylistic attributes. This allows us to train a powerful style representation that is independent of semantic content, using a contrastive learning objective.
    \item Multi-Condition Text Rendering: To ensure robust text generation even when replacing short words with long phrases (e.g., translation tasks), we incorporate multiple semantic signals into a custom architecture, combining rendered text glyphs with a character-level encoder.
\end{enumerate}

We evaluate STEPS against state-of-the-art methods including MOSTEL, TextDiffuser2, and TextCtrl. Our experiments demonstrate that STEPS achieves superior performance in style preservation and subjective quality while effectively handling complex translation tasks that require accommodating different text lengths.

\section{Related Work}
\label{sec:related_work}

Historically, GAN-based methods like SRNet and MOSTEL dominated STE. However, their reliance on isolated text patches precludes global image context, often necessitating complex blending steps that introduce visual artifacts.

Recent advances have shifted toward diffusion-based architectures. TextDiffuser2 \cite{chen_textdiffuser-2_2023} employs an inpainting approach to tackle visual text generation and editing jointly, offering state-of-the-art performance in readability and output quality. Alternatively, TextCtrl \cite{zeng2024textctrl} utilizes a diffusion-based approach with glyph-adaptive self-attention to disentangle style and structure features effectively at the patch level. More recently, RS-STE \cite{rsste} introduced a unified transformer framework that predicts text content and stylized images in parallel, employing a cyclic self-supervised fine-tuning strategy to disentangle style and content without paired training data.

We focus our comparisons on models with comparable scale and architectural goals. Therefore, we exclude FLUX-Text \cite{fluxtext}, which, despite achieving state-of-the-art multilingual performance via lightweight glyph modules, utilizes a 12-billion parameter model substantially larger than the $\sim$1B parameter range of our work. Similarly, we exclude AnyText \cite{tuo_anytext_2024} as initial experiments indicated its output quality was systematically lower than the aforementioned baselines.

Closest to our architectural approach is DiffUTE \cite{chen_diffute_2023}, which adapts Stable Diffusion 2 \cite{rombach2021highresolution} by replacing the text condition with a rendered target text image and swapping the CLIP encoder for TrOCR \cite{li_trocr_2022}. However, DiffUTE discards the input patch, forcing it to infer style solely from surrounding text; this limits its efficacy to documents with consistent typography (e.g., receipts). Our model overcomes this limitation by enriching TrOCR encodings with character-level and instruction embeddings. Furthermore, by incorporating a dedicated style encoder, our method recovers original text style even in the absence of surrounding stylistic cues.

\section{Method}
\label{section:method}

We propose a novel diffusion model architecture specialized in style-preserving scene text editing. Training only requires images and the corresponding extracted text and bounding boxes. Textual content is usually obtained with OCR engines such as PaddleOCR.

The architecture is based on Stable Diffusion 2 (SD2) inpainting \cite{rombach2021highresolution}, to which we add three input conditions. In this setting, the original image is encoded using a pre-trained Variational Auto-Encoder (VAE). A Gaussian perturbation is added to the encoded image, and the resulting noisy image is concatenated channel-wise with the encoded masked image and the (resized) binary mask. The resulting tensor is used as input to a U-Net denoising model that predicts noise in the latent space. Following the SD2 architecture, conditions are integrated via cross-attention mechanisms in the denoising model (Fig.~\ref{fig:cross_attention}). STEPS uses  one context vector for each of the four following conditions:

\begin{figure}[hbt!]
    \centering
    \includegraphics[width=\linewidth]{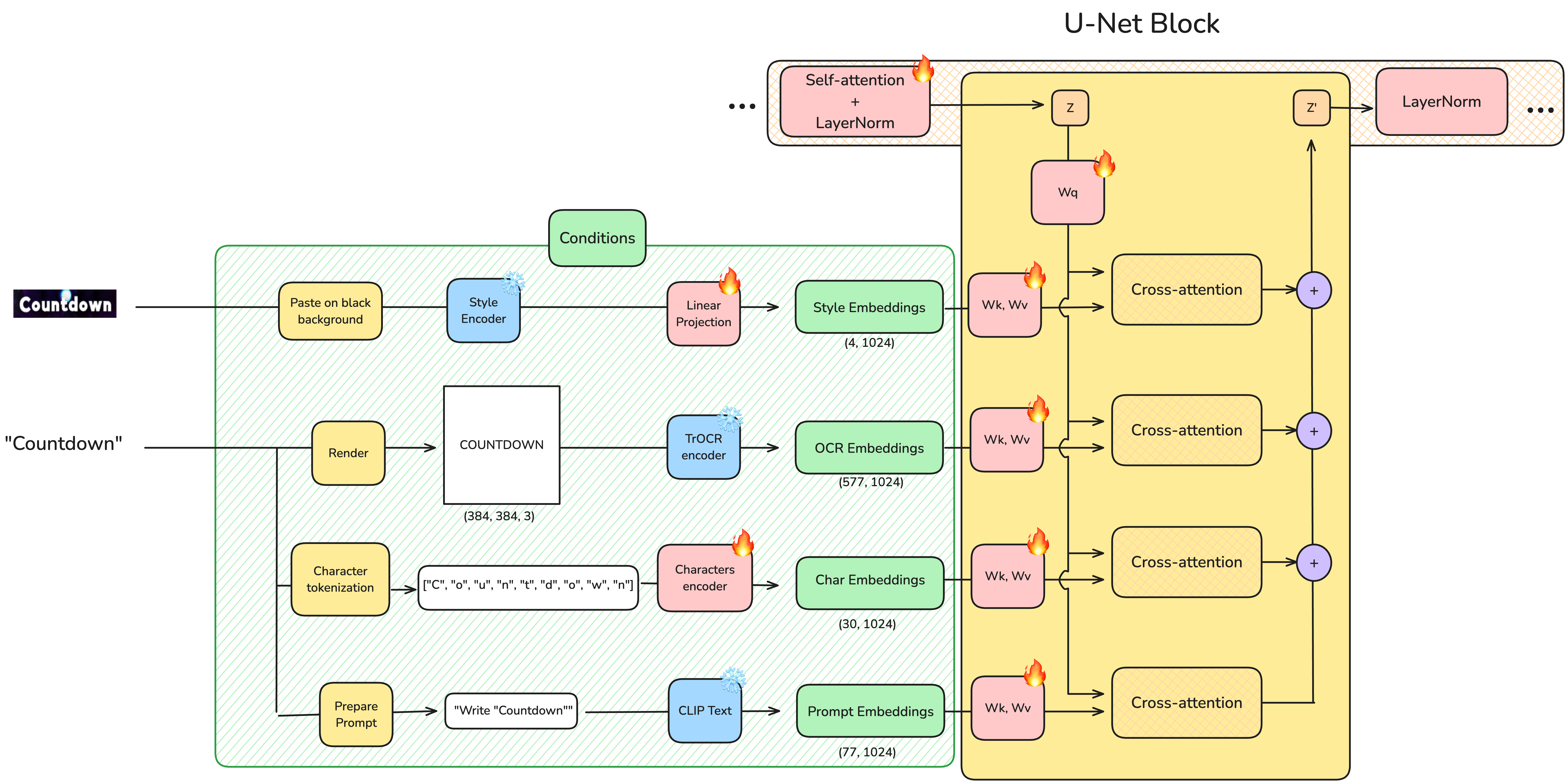}
    \caption{STEPS multi-conditions cross-attention module. From the source text patch and desired rendered text on the left, we compute four conditioning input embeddings (Style, OCR, Char, and Prompt). Each cross-attention layer of the U-Net is replaced by the sum of four corresponding cross-attention modules. The cross-attention modules share the same query vector $Q$ and their outputs are simply summed (details in the main text).}
    \label{fig:cross_attention}
\end{figure}

\textbf{Standard text prompt instruction}: we reuse the original text prompt condition from stable diffusion 2. For each target text \verb|target_text|, we simply pass the following string: \verb|write "target_text"| to the CLIP tokenizer and text encoder. It produces an embedding vector of size $(B, 77, 1024)$, where $B$ is the batch size.

\textbf{Character level encoder}: Character awareness has been found to improve visual text generation \cite{liu_character-aware_2023}. Indeed, standard text prompts use subword tokenizers that lose character-level information. To give character awareness to our model, we include a character tokenizer and encoder that maps the characters of the target text to embeddings with shape $(B, 30, 1024)$. Those embeddings are initialized randomly and then learned jointly with the rest of the parameters during training.

\textbf{Rendered target text}: Similarly to DiffUTE \cite{chen_diffute_2023}, the target text is rendered on a white background with a generic international typeface (GoNotoKurrent-Regular). The rendered text is an RGB image with a shape of 384x384. We pass that image through the encoder of a TrOCR model\footnote{\url{https://huggingface.co/microsoft/trocr-large-printed}}, producing embeddings of size $(B, 577, 1024)$.

\textbf{Style encoding}: to learn the original style of the input patch (font, color, background, etc.) without leaking content information, we introduce style encodings with a narrow bottleneck. The style encoder itself is detailed in the next section.

Mathematically, we augment each existing cross-attention layer in the U-Net architecture with three additional independent cross-attention layers. For each context vector $x_c$ ($c\in \{\text{prompt}, \text{char}, \text {render}, \text{style}\}$) we compute the key $K_c = x_c W_{k,c}$, and value $V_c = x_c W_{v, c}$ matrices. Inspired by IP-adapters \cite{ye_ip-adapter_2023} we use a shared query vector across cross-attention layers, i.e. for a query feature $Z$ we compute a single query matrix $Q = ZW_q$, and the output of the multi-conditions cross-attention module is given by
\[
\mathbf{Z}'=\sum_c\operatorname{Softmax}\left(\frac{Q K^{\top}_c}{\sqrt{d}}\right) V_c
\]
where all $W$ matrices are subsequently trained (conversely to \cite{ye_ip-adapter_2023}). The $W_q$, $W_{k,\text{prompt}}$, and  $W_{v,\text{prompt}}$ matrices are initialized from the SD2 checkpoint. Similarly to \cite{ye_ip-adapter_2023} we find that training converges faster when we initialize the new conditions attention matrices with the original cross-attention layer weights (i.e., $W_{v,c}=W_{v,\text{prompt}}, \; W_{k,c}=W_{k,\text{prompt}} \; \forall c$). Fig.~\ref{fig:cross_attention} illustrates the multi-condition cross-attention module visually.

\subsection{Style Encoder}

The style encoder is a critical piece of our model architecture. Its goal is to encode stylistic features of the text box to be edited (font, color, background, etc.) without leaking semantic content.

\begin{figure}
    \centering
    \includegraphics[width=\linewidth]{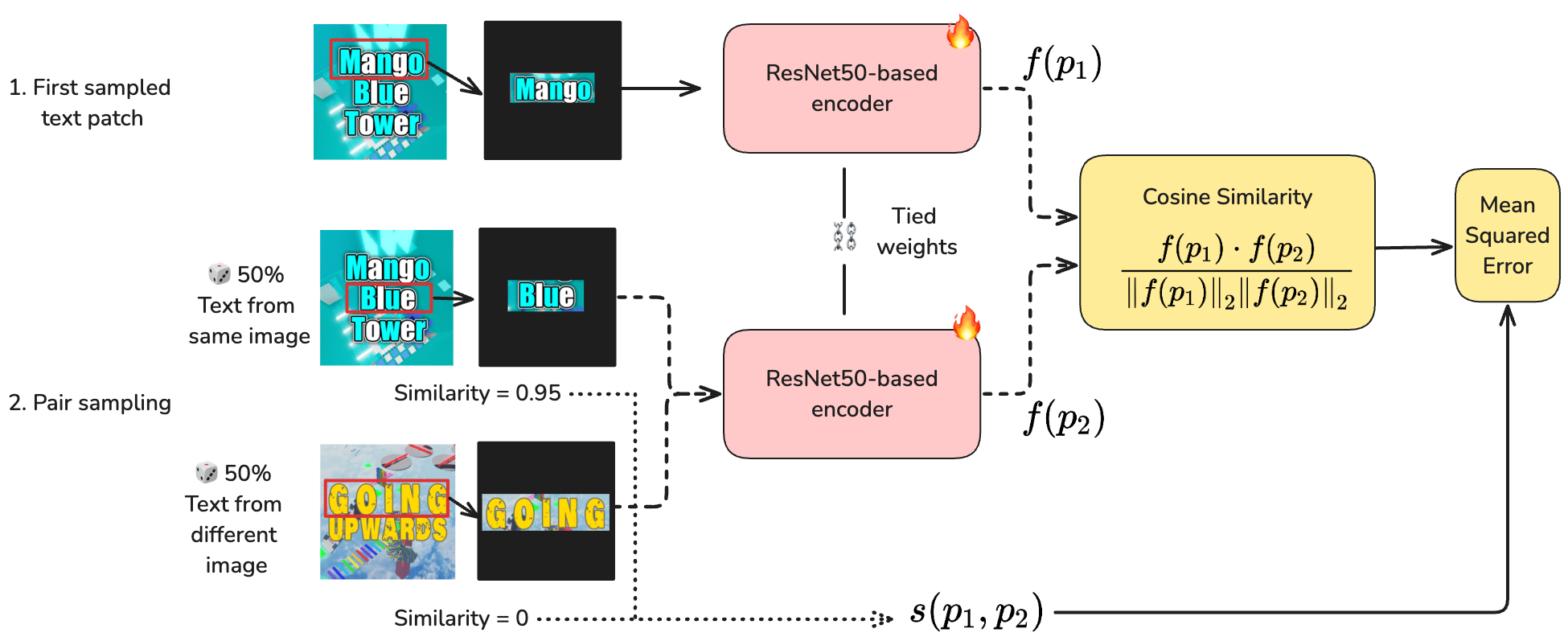}
    \caption{Style Encoder Training. To learn style embeddings (font, color, orientation, background) without encoding the textual content, we observe that stylistic similarity is correlated with spatial proximity in images that contain visual text. We then sample visual text patches from either different images (style similarity of zero) or the same image (style similarity of one minus bounding boxes distance). Finally, we train a bi-encoder on those stylistic similarity scores. A single copy of the resulting style encoder provides useful embeddings that we use as conditioning input for the main diffusion model. }
    \label{fig:style_encoder}
\end{figure}

We use a simple yet novel contrastive learning objective to train the style encoder, based on the following observation: in images containing text, text boxes close to one another tend to have a similar style because close words often belong to the same sentence and share similar backgrounds. We take advantage of this observed property to create a style similarity objective (Fig.~\ref{fig:style_encoder}).

From the AnyWord-3M dataset, we sample text box patches from either the same image or a different one (probability 50\%). Style similarity labels are generated for each patch pair $(p_1, p_2)$ by assigning a similarity of zero to patches from different images, and one minus the normalized bounding boxes distance if they are from the same image.

The image patches are then inpainted onto a 256x256 black background (and cropped if necessary). We then fine-tune a pre-trained ResNet50 model (pre-trained on ImageNet, with the last fully connected layer replaced with a randomly initialized 2048x1024 linear layer) in a bi-encoder setting. The loss function is the Mean-Squared Error between the labels described above and the cosine similarity between the embedding from both arms of the bi-encoder. Mathematically, for a pair of visual text patches $p_1$ and $p_2$, the loss function reads
\begin{equation}
     \ell (s, p_1, p_2) = \left(\frac{f(p_1) \cdot f(p_2)}{\left\| f(p_1)\right\|_2 \left\| f(p_2)\right\|_2} - s(p_1, p_2) \right)^2,
\end{equation}
where $f$ is the ResNet50-based image encoder and $s\in[0, 1]$ is the similarity label,

\begin{equation}
\begin{aligned}
  s(p_1, p_2) &=
    \begin{cases}
       1 - \min\limits_{b_1 \in B_1, b_2 \in B_2} \dfrac{\| b_1-b_2\|_2}{\sqrt{w\cdot h}} & \text{if } p_1 \text{ and } p_2 \text{ from same image}, \\
       0 & \text{otherwise}.
    \end{cases}
\end{aligned}
\end{equation}
$B_1$ and $B_2$ are the bounding boxes.

As potential limitations, this simple heuristic for style similarity may create a noisy objective (because texts with different styles are often present in the same image) or fail to disentangle style from content (because spatially close words have related meanings). The experiments presented in section~\ref{section:experiments} show that those limitations are minimal.

\subsection{Training}

We train STEPS on the AnyWord-3M-LAION dataset\footnote{\url{https://modelscope.cn/datasets/iic/AnyWord-3M/summary}}, which consists of images containing visual text mostly in English, with OCR annotations (bounding boxes positions and content). All images are 512x512 RGB images sampled from the Internet and selected for the legibility of their textual content.

Our model has 1.6B parameters, 882M of which are trainable. We trained for 5 days (40 epochs) on 8 A100-80G NVIDIA while freezing the VAE, CLIP prompt encoder, and TrOCR encoder. The character encoder is a linear embedding layer (dimension 128) followed by a 2-layer Transformer Encoder with 4 attention heads
per layer.

\section{Experiments}
\label{section:experiments}

Our method learns scene text editing capabilities from images with OCR annotations, in a self-supervised setting. We first train and evaluate the style encoder described in section~\ref{section:method}, then use it as an input to train STEPS.

As discussed in Section~\ref{sec:related_work}, we identified MOSTEL \cite{mostel}, TextDiffuser2 \cite{chen_textdiffuser-2_2023}, and TextCtrl \cite{zeng2024textctrl} as relevant competing methods. Of the many available methods, these three offer competitive performance, have a moderate number of parameters ($\sim$ 1B parameters or less), and have accessible open-source implementations. They also provide a representative sample of recent approaches to STE: a GAN-based method (MOSTEL), a diffusion model operating at the image level (TextDiffuser2), and a diffusion model focused on text patches (TextCtrl).

\begin{figure}[t]
    \centering
    {\scriptsize
    \setlength{\tabcolsep}{1pt}
    \begin{tabular}{@{}c@{\hspace{3pt}}ccc@{\hspace{6pt}}ccc@{\hspace{6pt}}ccc@{}}
        & \multicolumn{3}{c@{\hspace{6pt}}}{Original repainting} & \multicolumn{3}{c@{\hspace{6pt}}}{Indonesian Translation} & \multicolumn{3}{c@{}}{Italian Translation} \\[2pt]
        \gridlabel{Input} &
        \gridimg{0.1\linewidth}{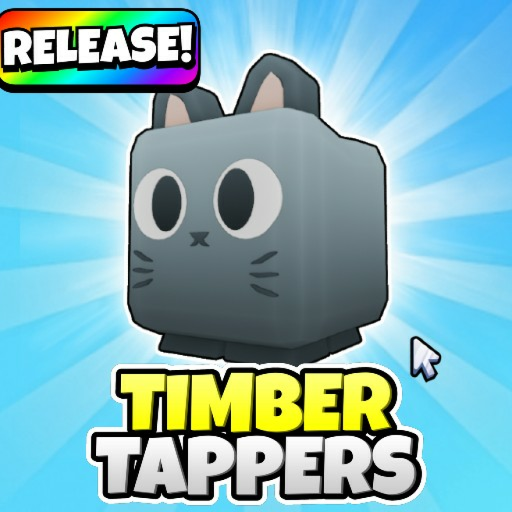} &
        \gridimg{0.1\linewidth}{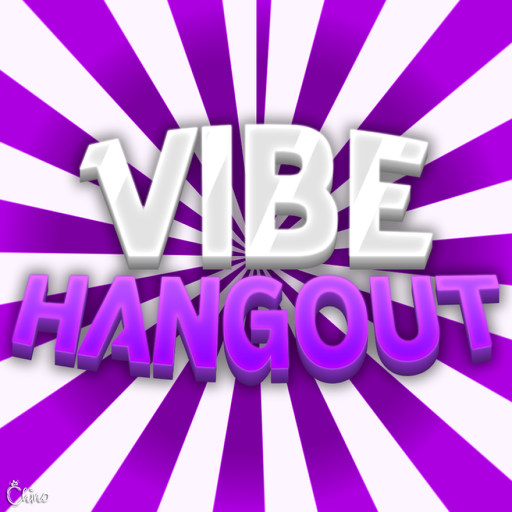} &
        \gridimg{0.1\linewidth}{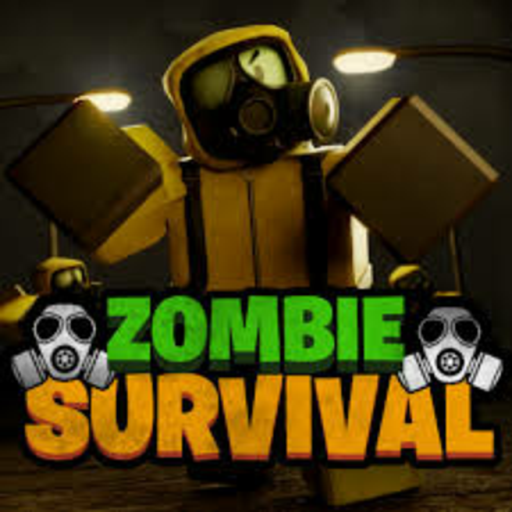} &
        \gridimg{0.1\linewidth}{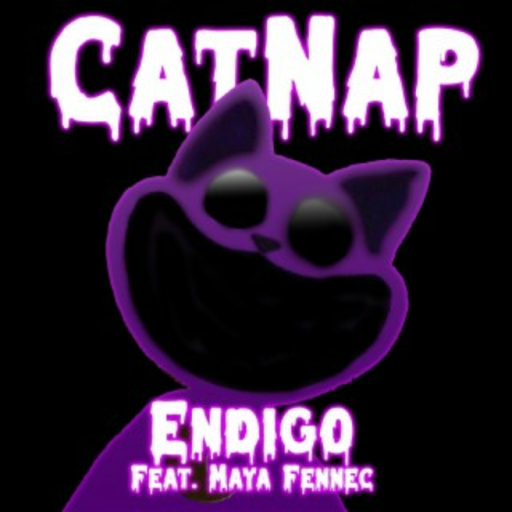} &
        \gridimg{0.1\linewidth}{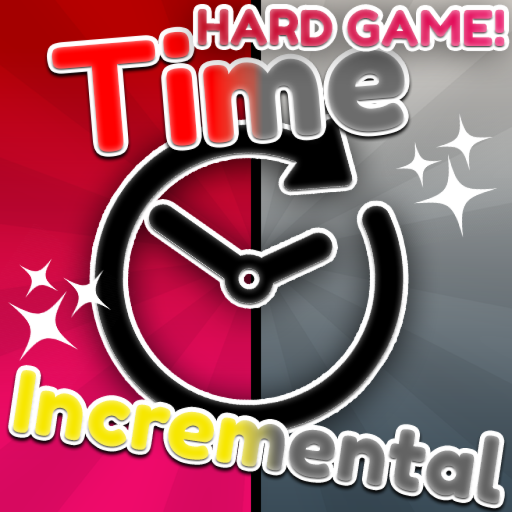} &
        \gridimg{0.1\linewidth}{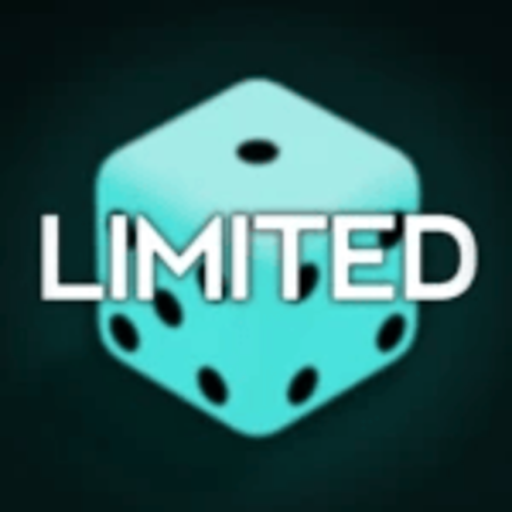} &
        \gridimg{0.1\linewidth}{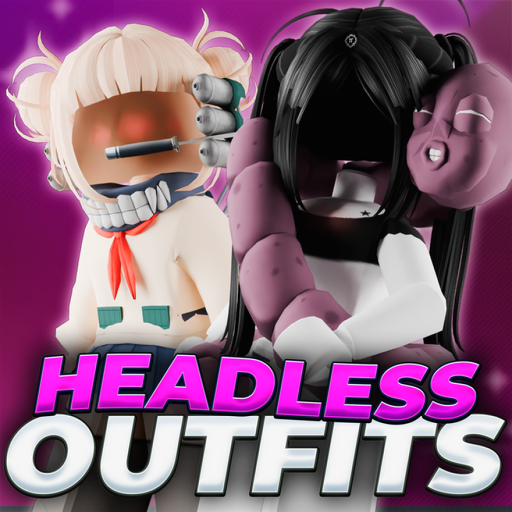} &
        \gridimg{0.1\linewidth}{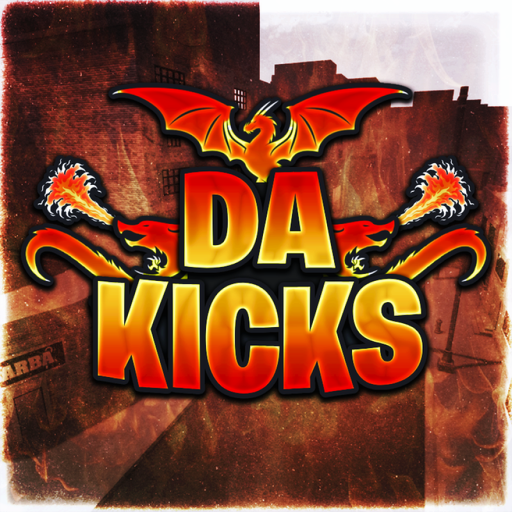} &
        \gridimg{0.1\linewidth}{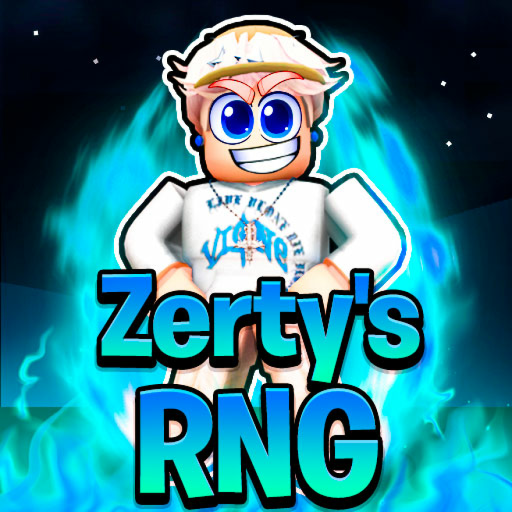} \\[2pt]
        \gridlabel{MOSTEL} &
        \gridimg{0.1\linewidth}{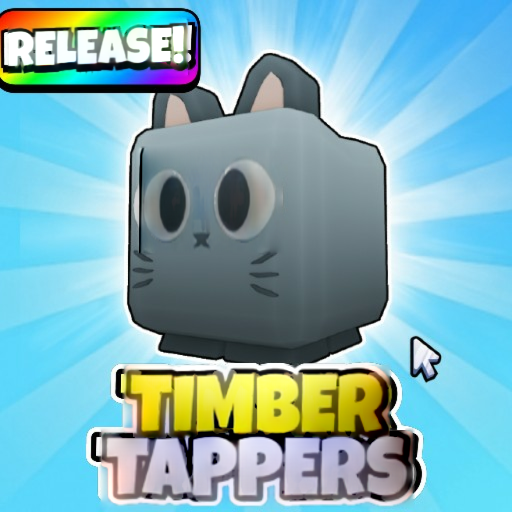} &
        \gridimg{0.1\linewidth}{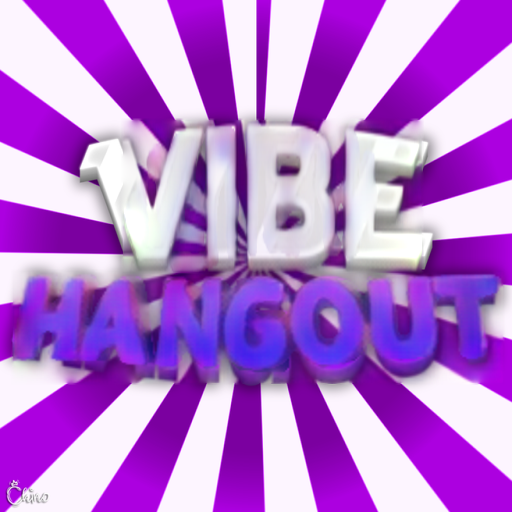} &
        \gridimg{0.1\linewidth}{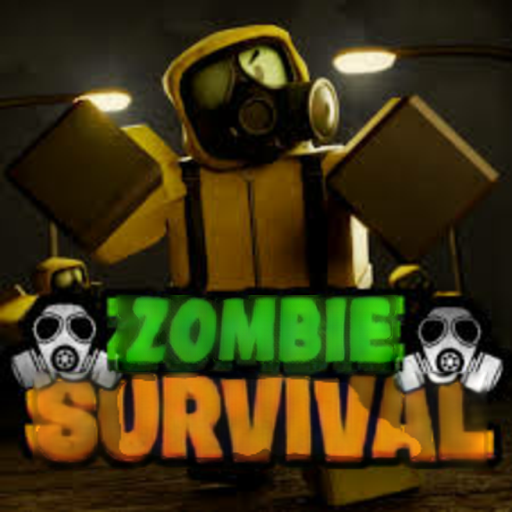} &
        \gridimg{0.1\linewidth}{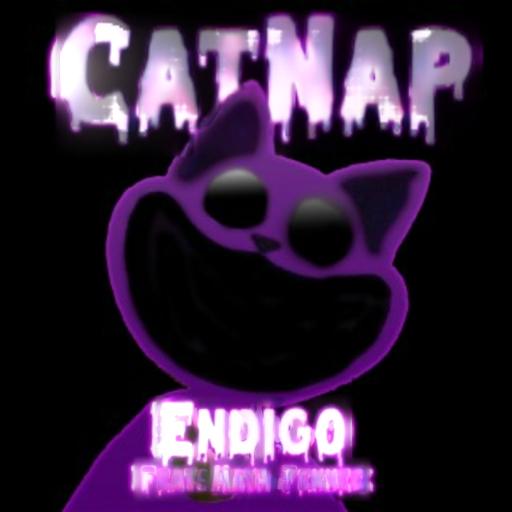} &
        \gridimg{0.1\linewidth}{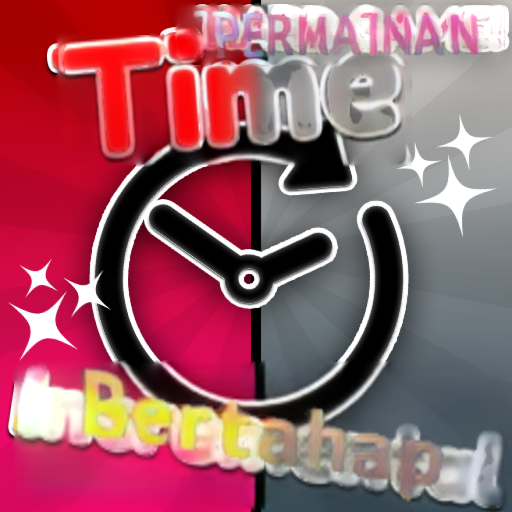} &
        \gridimg{0.1\linewidth}{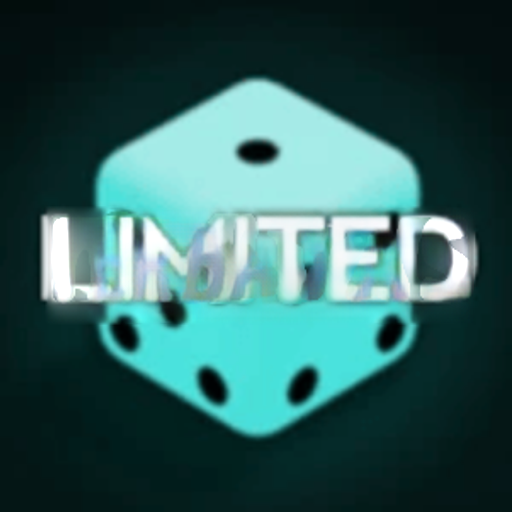} &
        \gridimg{0.1\linewidth}{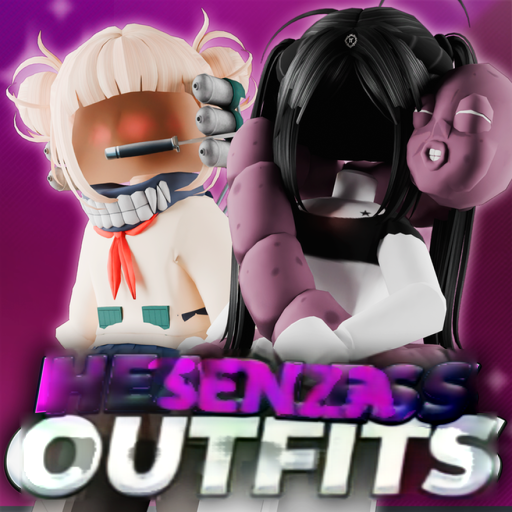} &
        \gridimg{0.1\linewidth}{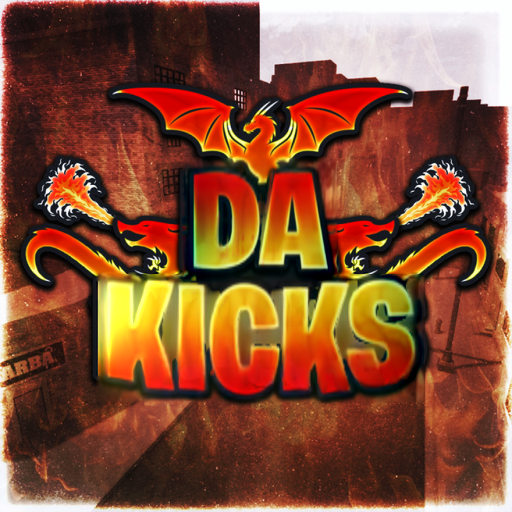} &
        \gridimg{0.1\linewidth}{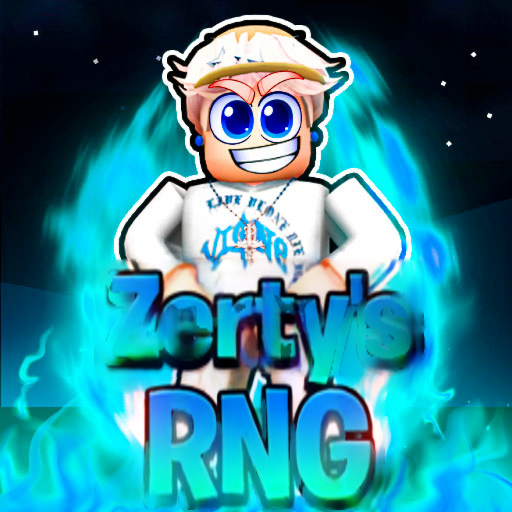} \\[2pt]
        \gridlabel{TextDiff. 2} &
        \gridimg{0.1\linewidth}{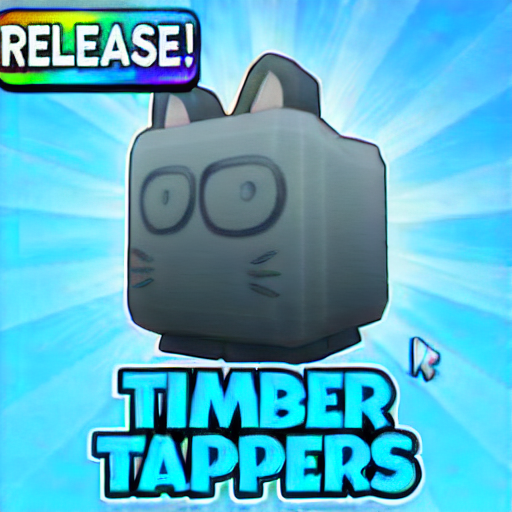} &
        \gridimg{0.1\linewidth}{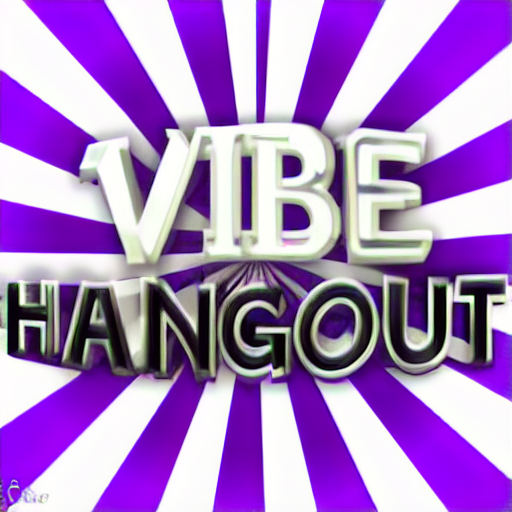} &
        \gridimg{0.1\linewidth}{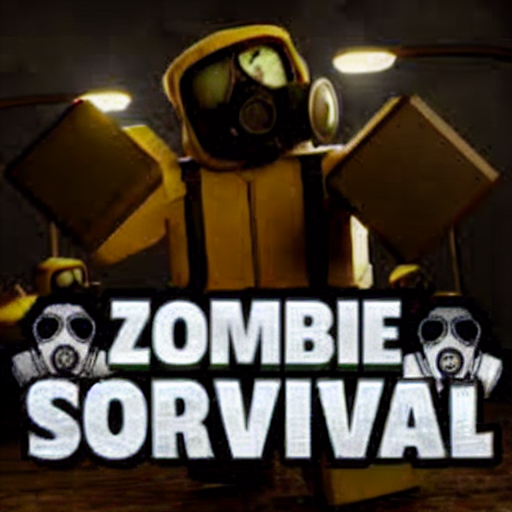} &
        \gridimg{0.1\linewidth}{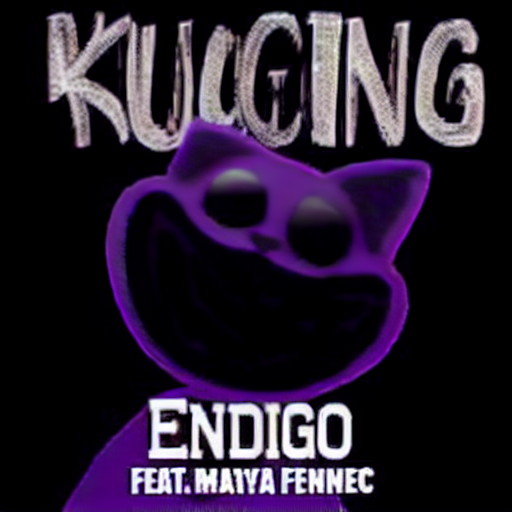} &
        \gridimg{0.1\linewidth}{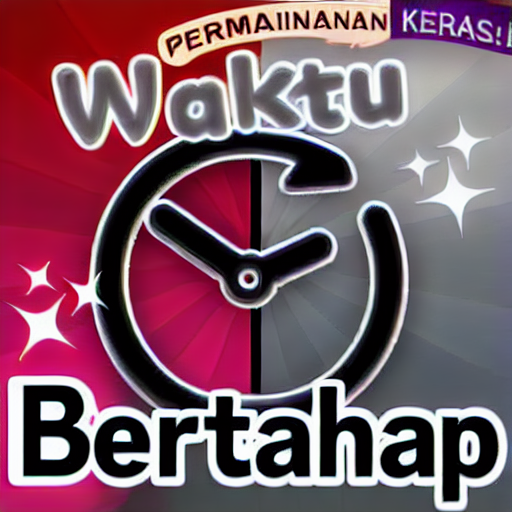} &
        \gridimg{0.1\linewidth}{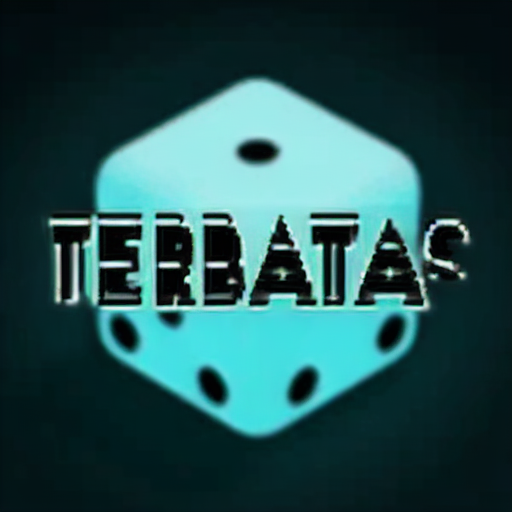} &
        \gridimg{0.1\linewidth}{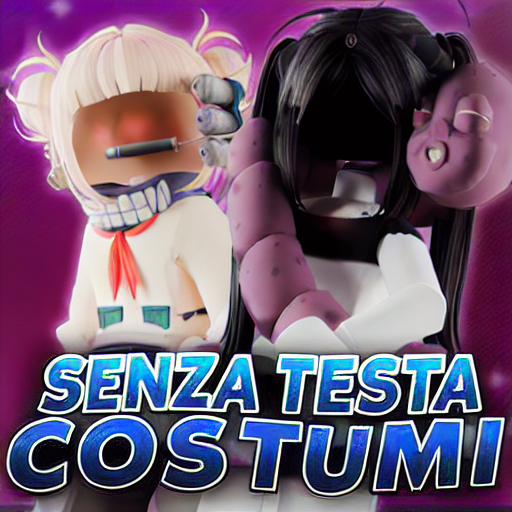} &
        \gridimg{0.1\linewidth}{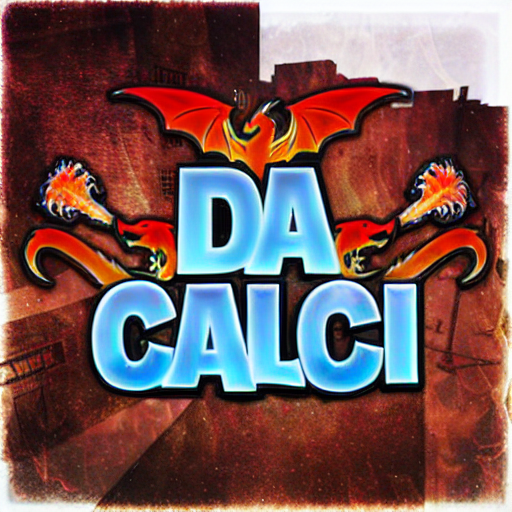} &
        \gridimg{0.1\linewidth}{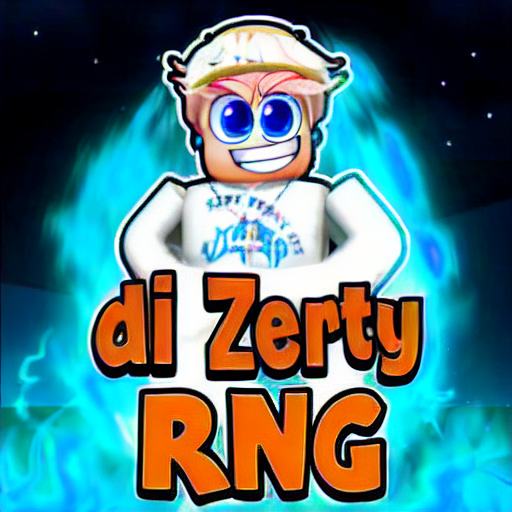} \\[2pt]
        \gridlabel{TextCtrl} &
        \gridimg{0.1\linewidth}{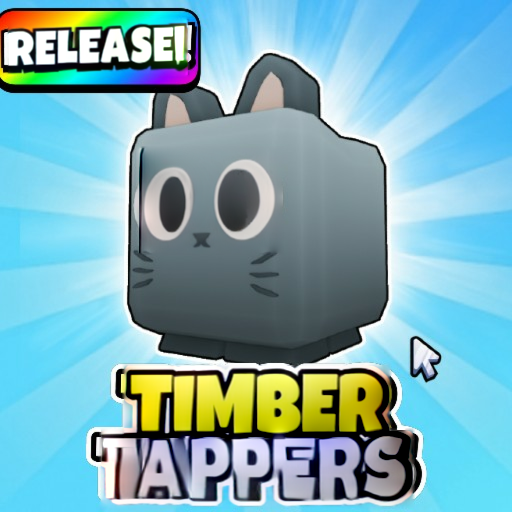} &
        \gridimg{0.1\linewidth}{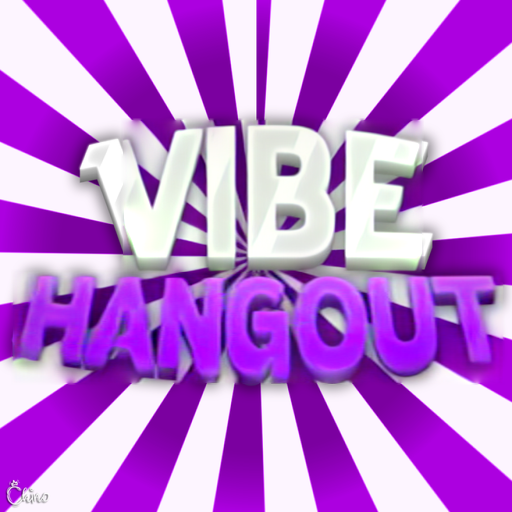} &
        \gridimg{0.1\linewidth}{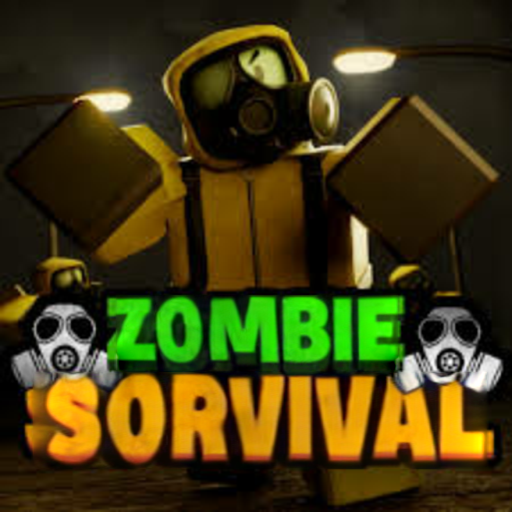} &
        \gridimg{0.1\linewidth}{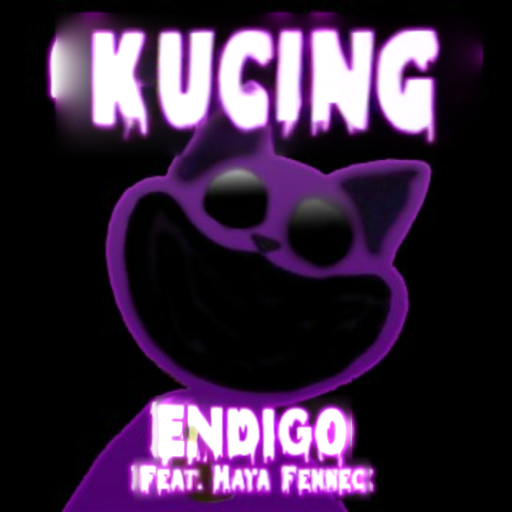} &
        \gridimg{0.1\linewidth}{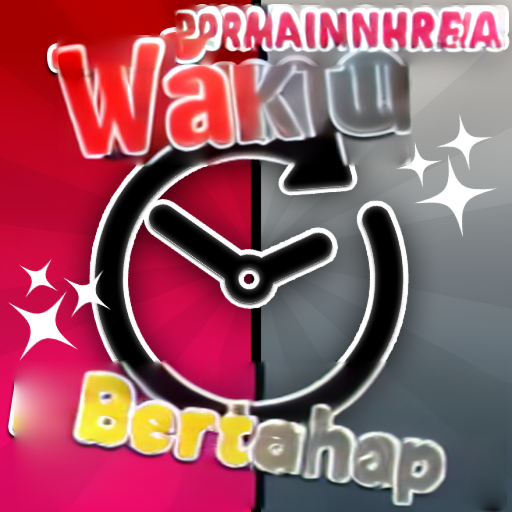} &
        \gridimg{0.1\linewidth}{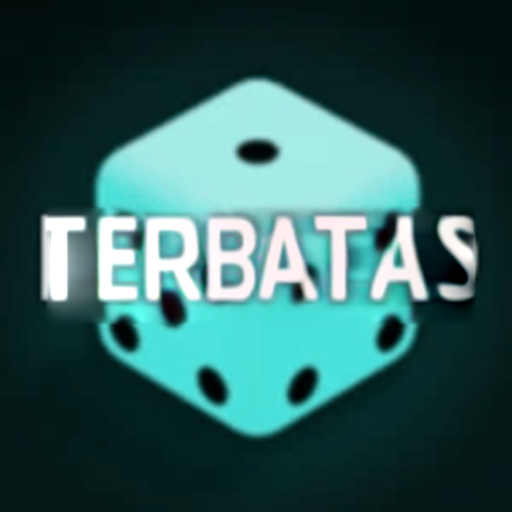} &
        \gridimg{0.1\linewidth}{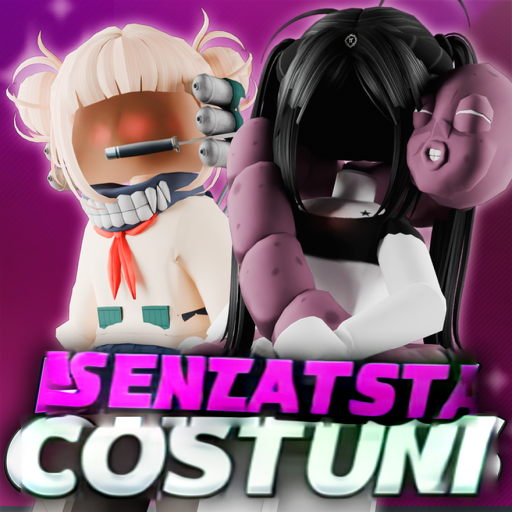} &
        \gridimg{0.1\linewidth}{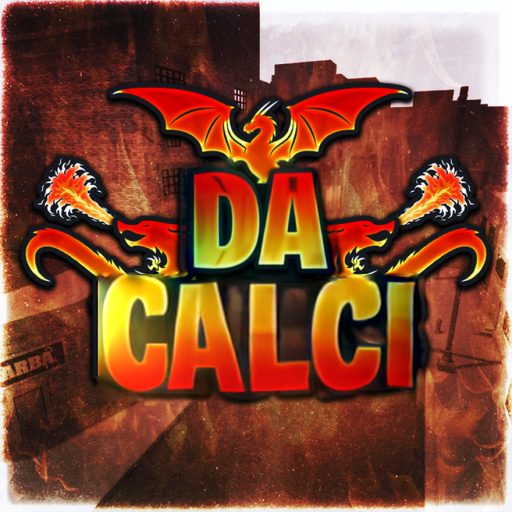} &
        \gridimg{0.1\linewidth}{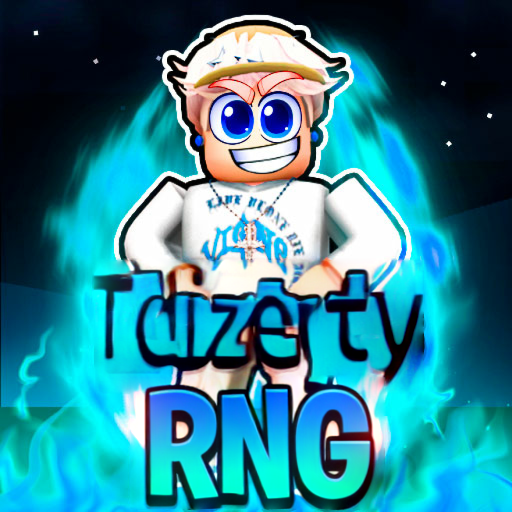} \\[2pt]
        \gridlabel{STEPS (new)} &
        \gridimg{0.1\linewidth}{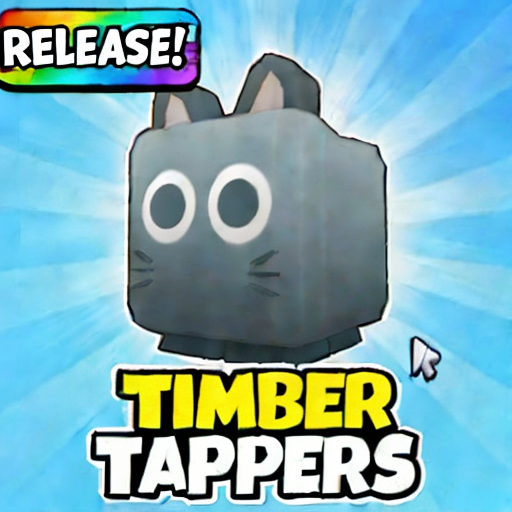} &
        \gridimg{0.1\linewidth}{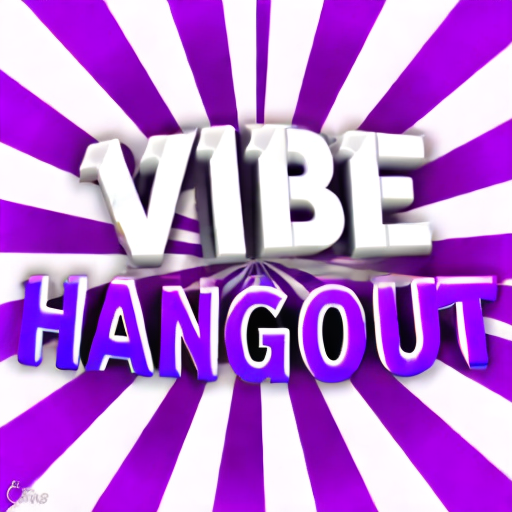} &
        \gridimg{0.1\linewidth}{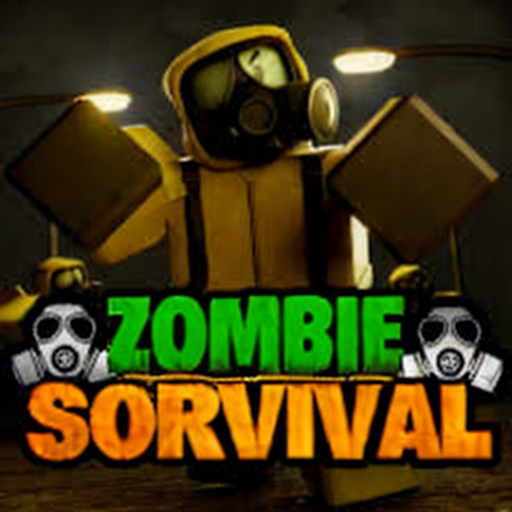} &
        \gridimg{0.1\linewidth}{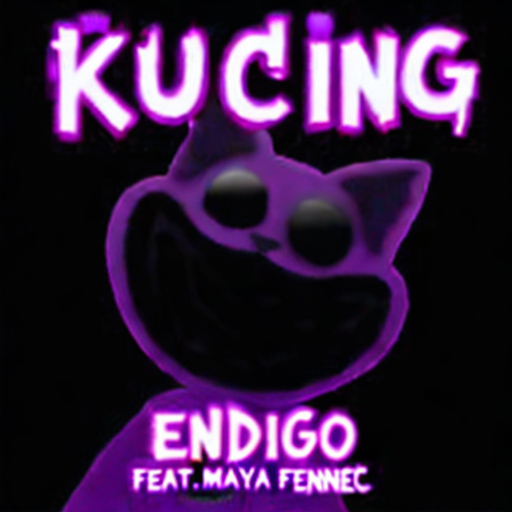} &
        \gridimg{0.1\linewidth}{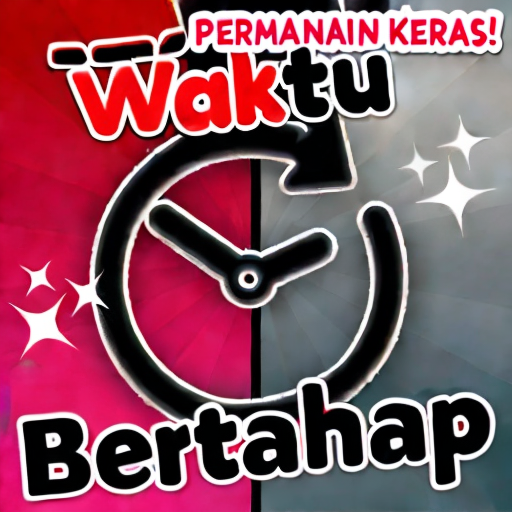} &
        \gridimg{0.1\linewidth}{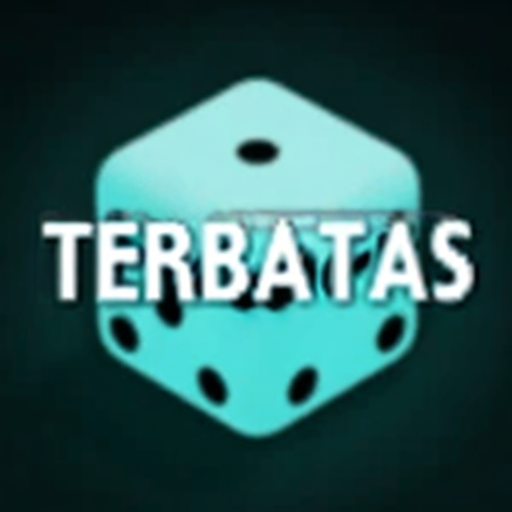} &
        \gridimg{0.1\linewidth}{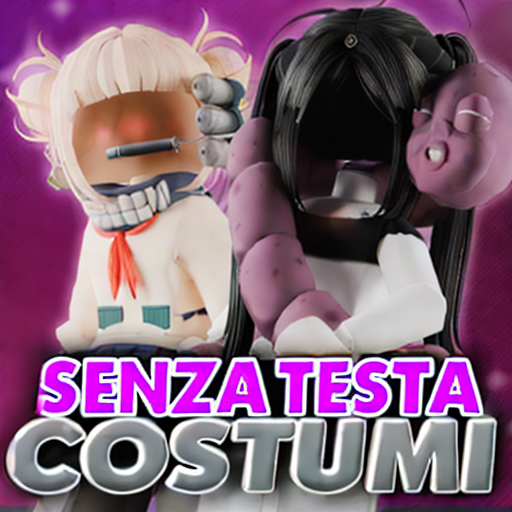} &
        \gridimg{0.1\linewidth}{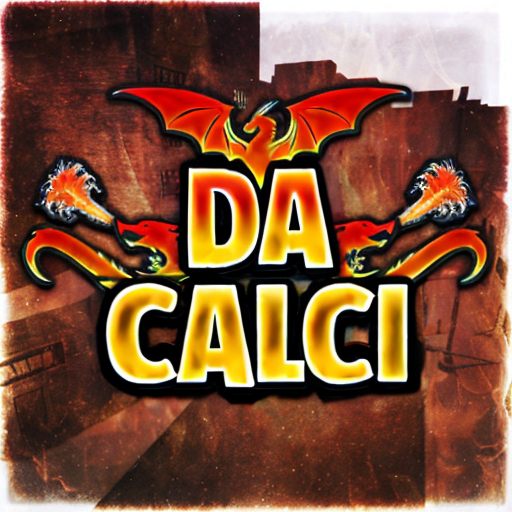} &
        \gridimg{0.1\linewidth}{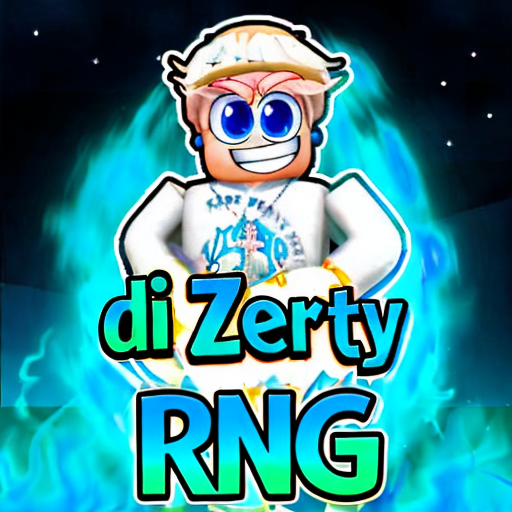} \\
    \end{tabular}}
    \caption{Samples comparison (best viewed zoomed in). In some of those examples, MOSTEL fails to properly erase the original text, while TextDiffuser only partially matches the style of the input text. TextCtrl preserves style best for some examples, but sometimes suffers from blurriness or fails to produce the correct letters. Our new STEPS method offers the best readability and style-preservation properties.}
    \label{fig:examples_comparison}
\end{figure}

\subsection{Style Encoder}

The Style Encoder is trained for two epochs on the AnyWord-3M dataset using a single NVIDIA A100-80G GPU, a process that takes roughly 48 hours. We allocate 5\% of the data for validation. After training, the model achieves a validation loss of 0.0315, compared to 0.4786 for a randomly initialized baseline evaluated on the same validation split.

\begin{table}[hbt!]
    \caption{Style encoder accuracy on the ScenePair retrieval task. To validate the training method of our style encoder, we treat pair retrieval within all 1,280 examples as a classification task. We perform nearest neighbor classification in the embedding space of the SigLIP2-base image encoder (86M parameters) and our style encoder (22M parameters). While being smaller, our model offers significantly higher accuracy, confirming that it captures style similarity more accurately than a generic image encoder such as SigLIP2.}
    \label{table:style_enc_qualitative}
    \centering
    \small
    \begin{tabular}{lcc}
        \toprule
        & Accuracy @ 1 & Accuracy @ 10 \\
        \midrule
        SigLIP2-base        & 0.1793          & 0.5354          \\
        STEPS Style Encoder & \textbf{0.5880} & \textbf{0.8882} \\
        \bottomrule
    \end{tabular}
\end{table}

To validate the heuristic we use to train the style encoder, we use the ScenePair dataset. While the ScenePair dataset contains only short and clean text in English, it contains 1,280 pairs of visual text patches that are stylistically identical but semantically distinct. In table~\ref{table:style_enc_qualitative} we compare pair retrieval with our style encoder and with the SigLIP2 image encoder.

\begin{figure}[ht]
    \centering
    \includegraphics[width=0.8\linewidth]{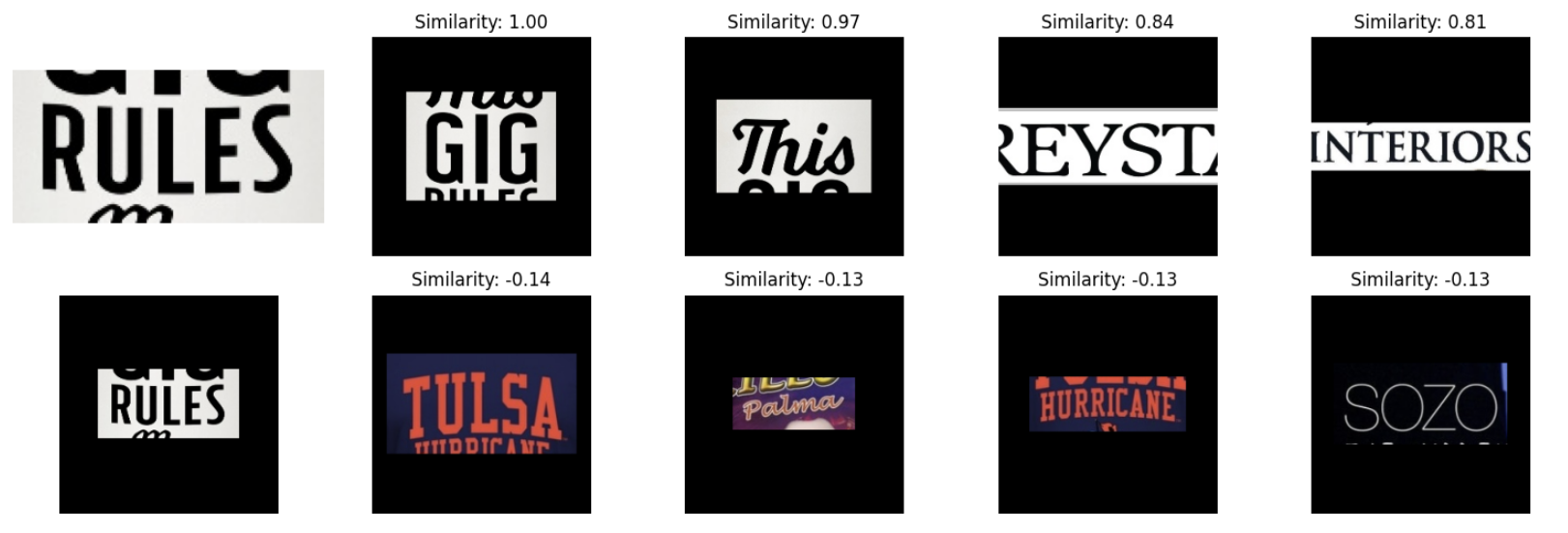}
    \caption{Style Encoder example for illustration purposes. The leftmost image is a sample text patch (top: original, bottom: preprocessed input to the encoding model), for which we computed the cosine similarity with all text patches in a sample of 1,000 images. The top row contains the most similar patches within this sample according to our style encoder, while the bottom row shows the least similar images. Text patches with similar colors, fonts, and backgrounds have aligned embeddings, while low embedding alignment correlates with text style discrepancies.}
    \label{fig:style-encoder-examples}
\end{figure}

Fig.~\ref{fig:style-encoder-examples} shows examples of an input text patch, together with its most similar and dissimilar examples in the validation set.
This example visually supports the idea that the style encoder, trained on spatial proximity as a proxy for style similarity, produces a representation that encodes visual text stylistic features, notably the text and background colors.

Those results support the assumption that spatial proximity correlates with style similarity in natural images, sufficiently to encode visual text style.

\subsection{Dataset}

Existing benchmarks (e.g., ScenePair, Anytext-benchmark) predominantly feature English text with matching source and target lengths. Consequently, they fail to adequately probe artifacts arising from significant length deviations or creative styling. We therefore evaluate our models using a translation task on the AnyWord-3M dataset. Following initial reconstruction tests on 1,000 annotated images, we translate text into Indonesian and Italian. These languages were selected to accommodate the English-only training of prior art (MOSTEL, TextDiffuser2, TextCtrl) while introducing necessary length variations.

The AnyWord-3M dataset includes 1,000 evaluation images with OCR annotations. We first test all four models on reconstructing images with masked OCR regions. Then, we translate the text into Indonesian and Italian---chosen for their English-subset alphabets, because prior art of MOSTEL, TextDiffuser2 and TextCtrl were trained only on English. This evaluates the models' ability to visually replace text of varying lengths.

\begin{table}[hbt!]
    \caption{Quantitative results on the AnyWord-3M-LAION benchmark dataset. Original refers to repainting the original image after masking text boxes, while Indonesian and Italian translations were selected because their alphabet is a subset of the English alphabet on which the models considered were trained. The translation task notably measures the ability to edit text that deviates from the original in length. Higher OCR indicates better readability, lower LPIPS better style preservation, and lower FID better output realism. Both MOSTEL and TextCtrl operate at the patch level. On the translation task, TextCtrl struggles with style consistency, while MOSTEL achieves strong style preservation---though often at the expense of readability, as it sometimes fails to fully erase the original text.}
    \label{table:results}
    \centering
    \small
    \setlength{\tabcolsep}{3.5pt}
    \begin{tabular}{@{}lccccccccc@{}}
        \toprule
        & \multicolumn{3}{c}{Original} & \multicolumn{3}{c}{Indonesian Translation} & \multicolumn{3}{c}{Italian Translation} \\
        \cmidrule(lr){2-4} \cmidrule(lr){5-7} \cmidrule(l){8-10}
        & LPIPS $\downarrow$ & FID $\downarrow$ & OCR $\uparrow$ & LPIPS $\downarrow$ & FID $\downarrow$ & OCR $\uparrow$ & LPIPS $\downarrow$ & FID $\downarrow$ & OCR $\uparrow$ \\
        \midrule
        MOSTEL        & 0.063 & 16.119 & 0.681 & 0.072 & 18.445 & 0.506 & 0.071 & 18.179 & 0.536 \\
        TextDiffuser2 & 0.080 & 16.095 & 0.690 & 0.099 & 19.226 & 0.651 & 0.097 & 18.879 & 0.653 \\
        TextCtrl      & 0.097 & 22.975 & 0.623 & 0.104 & 23.336 & 0.619 & 0.104 & 23.786 & 0.636 \\
        STEPS (new)   & \textbf{0.045} & \textbf{9.116} & \textbf{0.708} & \textbf{0.070} & \textbf{12.392} & \textbf{0.686} & \textbf{0.069} & \textbf{11.984} & \textbf{0.691} \\
        \bottomrule
    \end{tabular}
\end{table}

\subsection{Metrics}

We evaluate three aspects of scene text editing methods in the absence of reference outputs: Readability ((1 - normalized edit distance) between the target text and characters extracted by PP-OCRv4), Style-preservation (LPIPS to measure aesthetic consistency between the original and edited images), and Output quality (Fréchet Inception Distance to assess overall aesthetic realism).

All the methods we consider edit text boxes one at a time (although TextDiffuser 2 and STEPS take into account the entire image). For images that contain multiple text boxes, we experimented with several techniques and observe the best results when editing the images recursively for consistency and then blending all of the text boxes into the original image, using Poisson blending.

\begin{table}[hbt!]
    \caption{Vision-Language Model evaluation. In this setting of scene text translation with multiple text boxes per image, methods that operate at the text patch-level struggle with visual quality and legibility, from the perspective of the VLM evaluator. This is partly due to the blending post-processing step that is necessary to edit the entire input image. Inpainting methods like TextDiffuser2 and STEPS do not suffer from this limitation, and STEPS outperforms competing methods in both visual quality and legibility.}
    \label{table:results_vlm}
    \centering
    \small
    \newcommand{\twoline}[2]{\begin{tabular}[b]{@{}c@{}}#1\\#2\end{tabular}}
    \begin{tabular}{lcccc}
        \toprule
        & \multicolumn{2}{c}{Indonesian Translation} & \multicolumn{2}{c}{Italian Translation} \\
        \cmidrule(lr){2-3} \cmidrule(lr){4-5}
        & \twoline{Visual Quality}{(0--5) $\uparrow$} & \twoline{Legibility}{(0--5) $\uparrow$} & \twoline{Visual Quality}{(0--5) $\uparrow$} & \twoline{Legibility}{(0--5) $\uparrow$} \\
        \midrule
        MOSTEL        & 1.52  & 1.38  & 1.84  & 1.68  \\
        TextDiffuser2 & 2.46  & 2.83  & 2.56  & 3.01  \\
        TextCtrl      & 1.496 & 1.536 & 1.424 & 1.472 \\
        STEPS (new)   & \textbf{3.224} & \textbf{3.624} & \textbf{3.544} & \textbf{3.936} \\
        \bottomrule
    \end{tabular}
\end{table}

\begin{table}[hbt!]
    \caption{Quantitative results from the ablation study. Higher OCR values indicate improved readability, while lower LPIPS scores reflect better style preservation, and lower FID scores suggest higher output realism. Removing the style encoder from the conditioning inputs negatively affects style preservation and realism, but has a more limited impact on readability.}
    \label{table:ablations}
    \centering
    \small
    \setlength{\tabcolsep}{3.5pt}
    \begin{tabular}{@{}lccccccccc@{}}
        \toprule
        & \multicolumn{3}{c}{Original} & \multicolumn{3}{c}{Indonesian Translation} & \multicolumn{3}{c}{Italian Translation} \\
        \cmidrule(lr){2-4} \cmidrule(lr){5-7} \cmidrule(l){8-10}
        & LPIPS $\downarrow$ & FID $\downarrow$ & OCR $\uparrow$ & LPIPS $\downarrow$ & FID $\downarrow$ & OCR $\uparrow$ & LPIPS $\downarrow$ & FID $\downarrow$ & OCR $\uparrow$ \\
        \midrule
        STEPS (all conditions) & \textbf{0.045} & \textbf{9.116} & \textbf{0.708} & \textbf{0.070} & \textbf{12.392} & \textbf{0.686} & \textbf{0.069} & \textbf{11.984} & \textbf{0.691} \\
        W/o style encoder      & 0.056 & 11.501 & 0.701 & 0.083 & 15.416 & 0.676 & 0.080 & 14.848 & 0.672 \\
        \bottomrule
    \end{tabular}
\end{table}

\subsection{Vision-Language Model evaluation}

Beyond the metrics reported in the previous section, we also assess the legibility and aesthetic quality of the images generated by different methods using a Vision-Language Model (VLM). For this evaluation, we employ GPT-4o as the VLM and ask it to rate visual quality and legibility from 0 to 5 each.

\subsection{Results}

The quantitative  (Tab.~\ref{table:results}) and VLM (Tab.~\ref{table:results_vlm}) evaluations show that STEPS matches or surpasses the performance of both state-of-the-art GAN-based models and leading diffusion-based approaches to text editing. Fig.~\ref{fig:examples_comparison} illustrates the differences with examples.

To preserve the original image parts far from text boxes, we use Blended Latent Diffusion \cite{avrahami_blended_2023} during denoising. This constraint enables recursive editing, so we can successively edit different text boxes in the same image. We use DPM-Solver \cite{lu_dpm-solver_2022} with 30 steps for the denoising schedule.

\subsection{Ablation Studies}

Our method adds three components to the underlying Stable Diffusion 2 inpainting model, through a multifold conditioning module (Fig.~\ref{fig:cross_attention}). The first component, the style encoder, captures the style of input text patches (font, color, background, etc.). The remaining two components (TrOCR encoder and character-level encoder) aim to improve the glyphs and character awareness of the model.

Since the introduction of the style encoder is our main contribution, we train a model without it and evaluate it against our full STEPS model (Table~\ref{table:ablations}).

For both versions, the model is retrained from the same initial weights and with the same compute budget (5 days on 8 A100-80G NVIDIA). The checkpoint with the lowest validation loss is used (on a subsample of 70k images).

Results from the ablation studies validate the expected contribution of the style encoder. Removing it degrades aesthetic quality and style consistency.

These findings alleviate concerns that the style encoder inadvertently captures semantic content. If the encoder were leaking substantial semantic information, we would expect the model to exhibit a disproportionate performance advantage in the Original repainting task. However, the observed performance differential between the Original repainting and translation tasks remains consistent regardless of the style encoder's inclusion. This indicates that the disparity is intrinsic to the training dataset rather than an artifact of semantic leakage from the style encoder.

\section{Conclusion}

We introduce STEPS, a diffusion-based visual text editing technique with style transfer, addressing a key limitation in prior STE methods through the introduction of a novel style-preserving encoder. Benchmarking against GAN and diffusion-based approaches, STEPS achieves strong style adherence, seamless text blending, and high legibility across diverse backgrounds and text lengths.

Despite its advancements, STEPS has limitations, such as occasional superfluous glyphs. Future work could enhance robustness through improved style encoder training, larger VAE models, or alternative backbones like diffusion transformers and flow matching.

{\small
\bibliographystyle{plainnat}
\bibliography{references}
}

\end{document}